\documentclass[conference]{IEEEtran}
\IEEEoverridecommandlockouts

\usepackage{cite}
\usepackage{amsmath,amssymb,amsfonts}
\usepackage{algorithmic}
\usepackage{graphicx}
\usepackage{textcomp}
\usepackage{xcolor}
\usepackage{url}
\def\BibTeX{{\rm B\kern-.05em{\sc i\kern-.025em b}\kern-.08em
    T\kern-.1667em\lower.7ex\hbox{E}\kern-.125emX}}
\begin{document}

\title{Industry 4.0 Asset Administration Shell (AAS): Interoperable Skill-Based Service-Robots
\thanks{This work-in-progress paper is supported by the German Federal Ministry for Economic Affairs and Energy (BMWi) in the programme "Development of Digital Technologies (PAiCE)" under grant agreement No. 01MA17003D, project {\it SeRoNet - Eine Plattform zur arbeitsteiligen Entwicklung von Serviceroboter-Lösungen} and by EFRE Program Baden-Württemberg 2014-2020 (ZAFH Intralogistik).}
}

\author{\IEEEauthorblockN{Vineet Nagrath}
\IEEEauthorblockA{\textit{Service Robotics} \\
\textit{Technische Hochschule Ulm}\\
89075 Ulm, Germany \\
Vineet.Nagrath@thu.de}
\and
\IEEEauthorblockN{Timo Blender}
\IEEEauthorblockA{\textit{Service Robotics} \\
\textit{Technische Hochschule Ulm}\\
89075 Ulm, Germany \\
Timo.Blender@thu.de}
\and
\IEEEauthorblockN{Nayabrasul Shaik}
\IEEEauthorblockA{\textit{Service Robotics} \\
\textit{Technische Hochschule Ulm}\\
89075 Ulm, Germany \\
Nayabrasul.Shaik@thu.de}
\and
\IEEEauthorblockN{Christian Schlegel}
\IEEEauthorblockA{\textit{Service Robotics} \\
\textit{Technische Hochschule Ulm}\\
89075 Ulm, Germany \\
Christian.Schlegel@thu.de}
}

\maketitle

\begin{abstract}
This paper describes our use of Industry 4.0 Asset Administration Shells (AASs) in the context of service robots. 
We use AASs with software components of service robots and with complete service robot systems. 

The AAS 
for a software component serves as a standardized digital data sheet. It helps sysem builders at design time in finding 
and selecting software components that match system-level requirements of the systems to be built. 

The AAS for a system 
comprises a data sheet for the system and furthermore collects at runtime operational data and it allows for skill-level 
commanding of the service robot. 

AASs are generated and filled as part of our model-driven development and composition 
workflow for service robotics. AASs can serve as a key enabler for a standardized integration and interaction with service robots.
\end{abstract}

\begin{IEEEkeywords}
service robotics, asset administration shell, skill-based, model-driven software development
\end{IEEEkeywords}

%
\section{Introduction}

\begin{figure}[htbp]
\centerline{\includegraphics[width=\columnwidth]{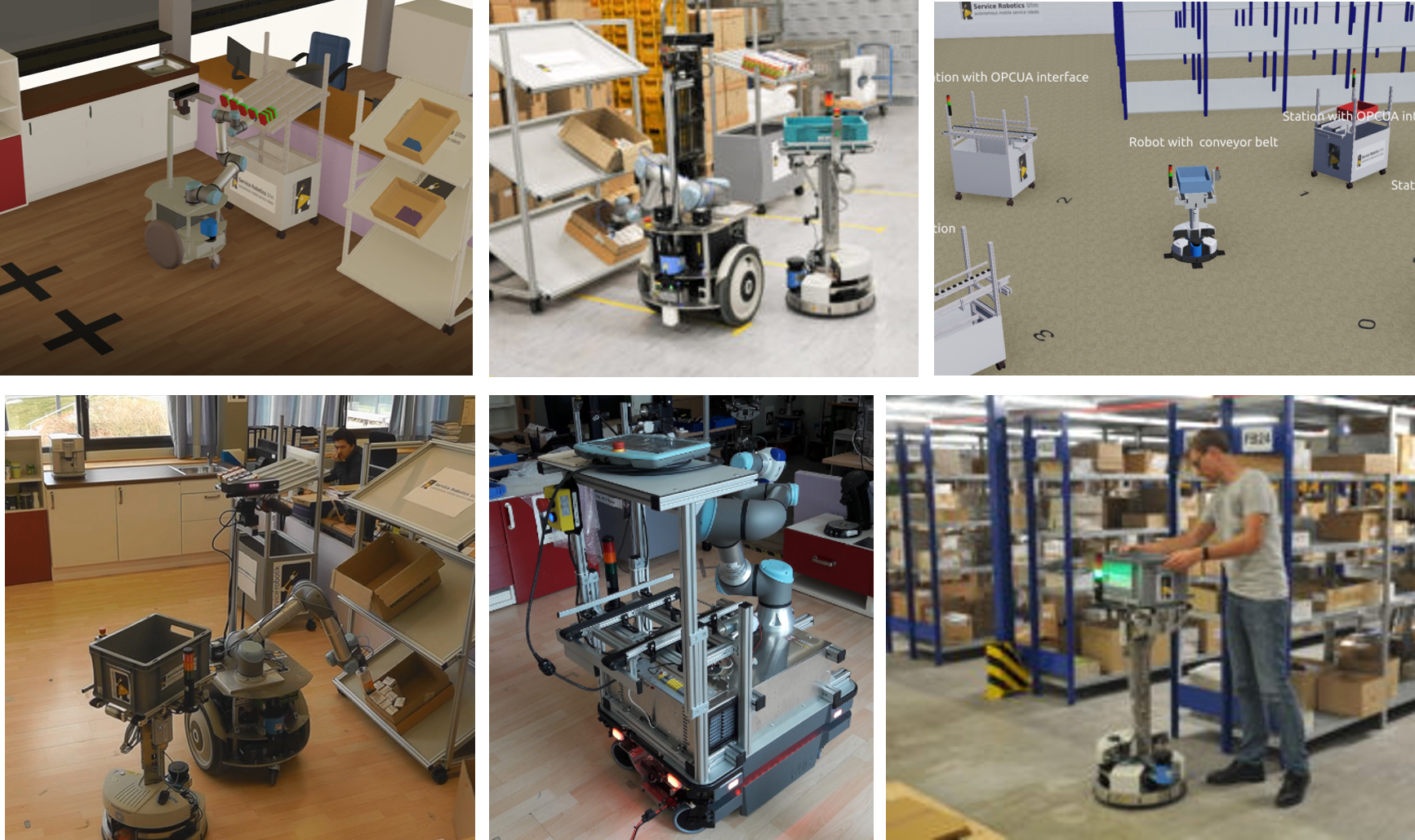}}
\caption{Our service robots for intralogistic tasks like transportation and order picking in simulation, in our lab environment and in relevant 
real-world settings. The robots are of type FESTO Robotino with a conveyor belt, Segway mobile platform with a UR5 manipulator (called {\em Larry}) and 
MiR 250 mobile robot with both, a UR5e manipulator and a conveyor belt (called {\em Macy}). The different types of robots are part of a mixed fleet offering 
different capabilities with respect to navigation and manipulation. Their software is built with {\sc SmartMDSD} and they can now be commanded via their AASs.}
\label{fig-robots}
\end{figure}

An advanced service robot is the promise of a multi-purpose flexible machine. It shall robustly fulfill tasks even 
in open-ended environments and in workspaces shared with e.g. persons. Users want to use a service robot as
an assistant that can be adapted with low effort to new tasks. Service robots are not isolated machines anymore 
but they have to interact with complex infrastructure and machinery (fig. \ref{fig-robots}). Identifying and selecting
service robots that match upcoming tasks is crucial as is the identification and selection of fitting software components
to equip a service robot with the required capabilities.

The AAS is one of the key concepts of Industry 4.0 to achieve flexibility and variability within factories and 
across value networks. An industry 4.0 Asset Administration Shell (AAS) describes an asset in a standardized 
manner. It envisions a standardized exchange of information about an asset and describes how to interact 
with it. Submodels structure information and one can present even the same information within different and 
co-existing submodels that all address a different domain.

At present many standardization efforts for submodels \cite{ZVEI-AAS-SM} and efforts for harmonizing AASs take place and a
bunch of new AAS submodel templates are under development \cite{idta}. Nevertheless, there is still a tremendous 
need for real-world examples that underpin what makes a reasonable scope and a good template for a concrete 
AAS submodel and which kind of submodels come with real benefits.

We extended {\sc SmartMDSD} \cite{smartmdsd} to support the generation of AASs for software components and service robot
systems, respectively. The implementation of the AAS uses the Eclipse BaSyx SDK \cite{basyx}. As first use-cases, the AAS allows at 
runtime to command a service robot at skill-level and to collect individual performance indicators. The gained insights now 
shape our next steps in extending our AAS submodels for service robots.

%
\section{The Service Robotics Business Ecosystem}

\begin{figure}[htbp]
\centerline{\includegraphics[width=\columnwidth]{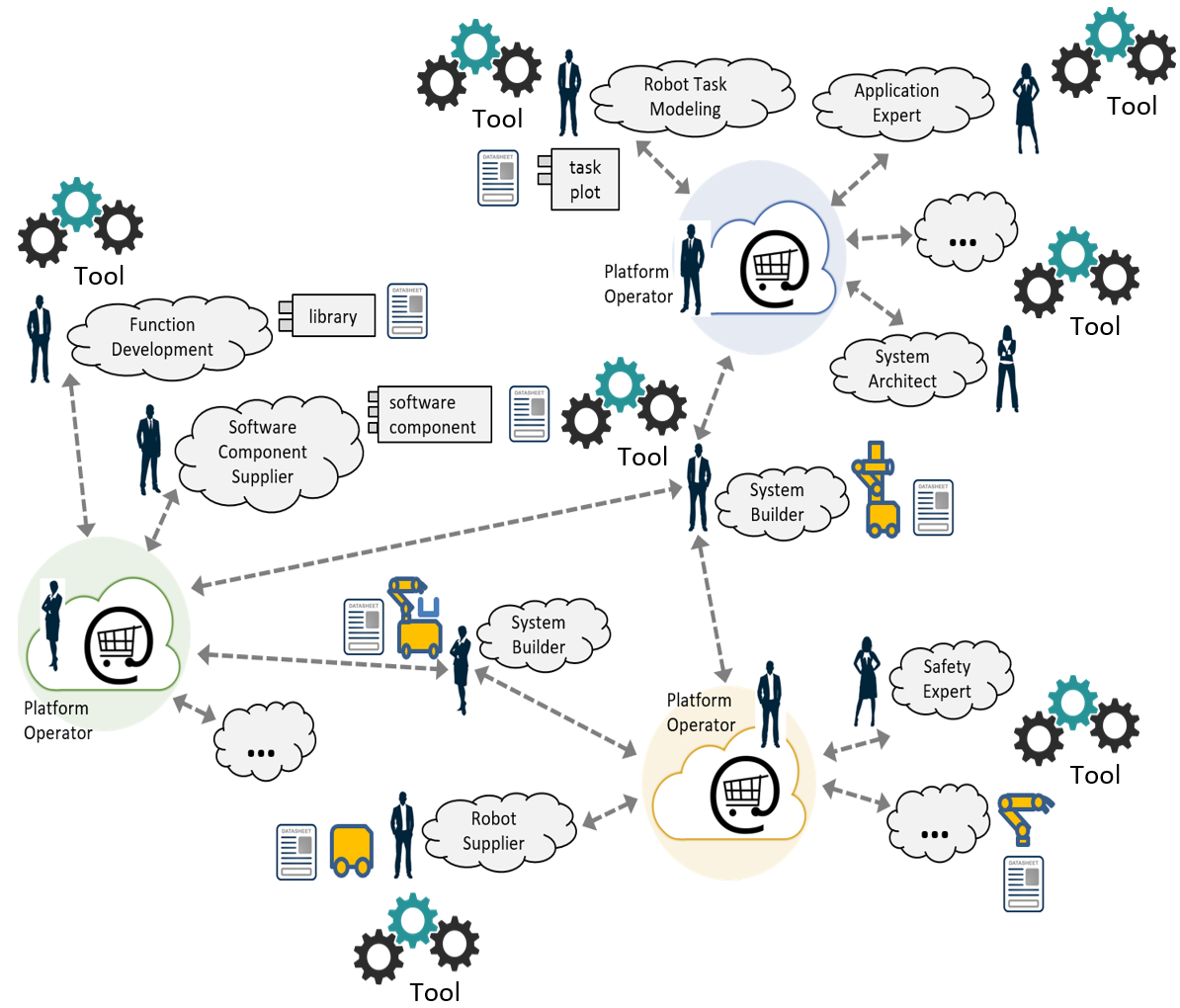}}
\caption{Overview on the service robotics ecosystem with different roles, assets with digital data sheets, tool support and market places. 
It is based on compsability, compositionality and separation of roles. The open-source and free-to-use Eclipse-based model-driven toolchain
{\sc SmartMDSD} is the easiest way to participate in the service robotics software business ecosystem as it provides 
role-specific support for the different ecosystem participants (such as component developers, system builders, and others).}
\label{fig-ecosystem}
\end{figure}

\begin{figure}[htbp]
\centerline{\includegraphics[width=\columnwidth]{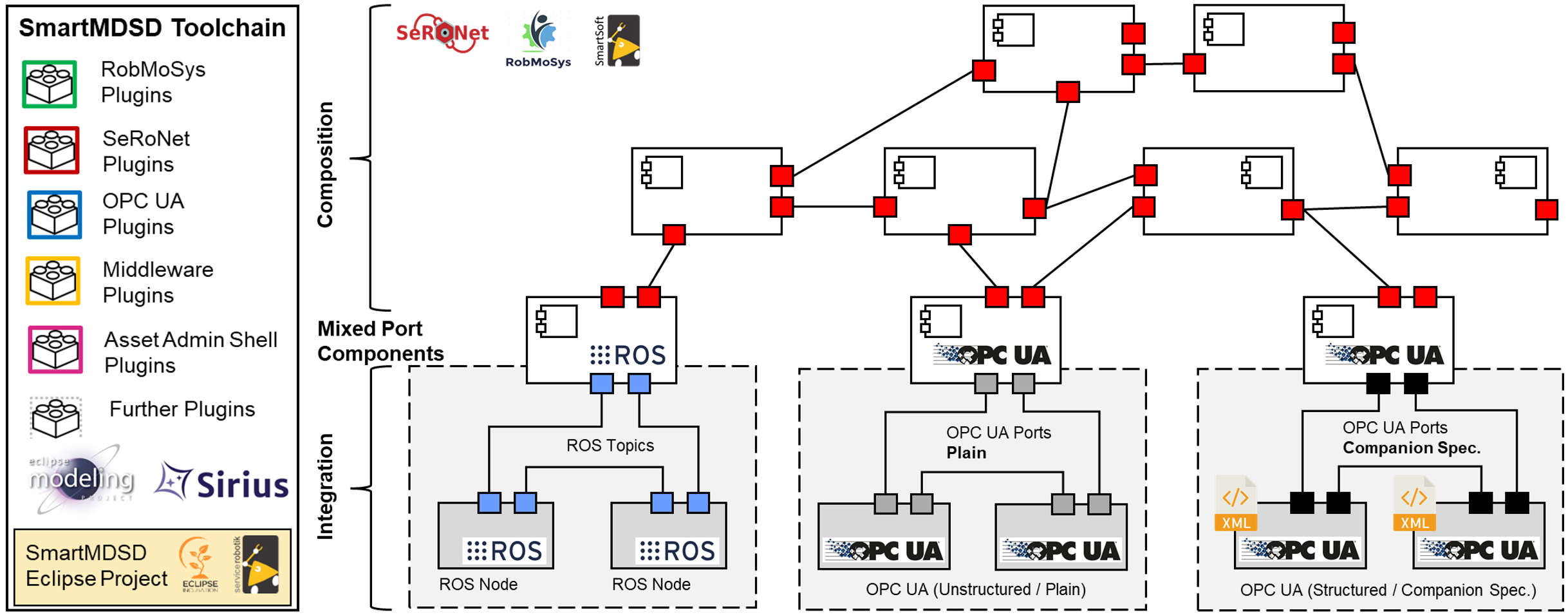}}
\caption{The toolchain {\sc SmartMDSD} can be extended by plugins (e.g. to support the generation of AASs). {\em Mixed-Port Components} 
interface between composable software components (indicated by the red ports: these fully adhere to the ecosystem composition structures) 
and other worlds (such as ROS, OPC UA, and others: these do not adhere to the ecosystem composition structures but there is a need to 
interface to them).}
\label{fig-components}
\end{figure}

\begin{figure}[htbp]
\centerline{\includegraphics[width=\columnwidth]{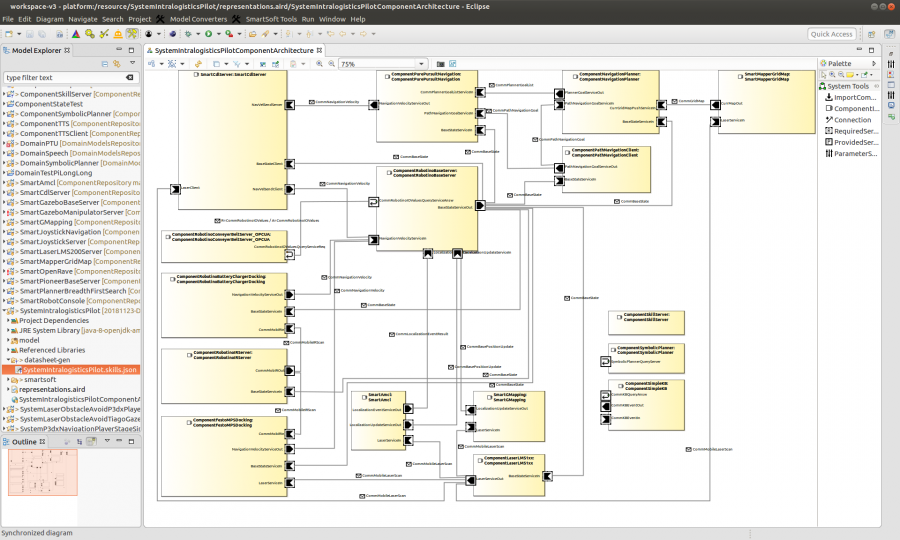}}
\caption{The view of the system builder in {\sc SmartMDSD}: component selection, component composition, adding task plots and performing deployment.}
\label{fig-eclipse}
\end{figure}

The business ecosystem for service robotic software components and systems now gains more and more momentum
\cite{Schlegel2021} \cite{robmosys} \cite{seronet} \cite{stewardship} \cite{xito}. Fig. \ref{fig-ecosystem} gives an 
overview on the service robotics business ecosystem with selected roles, assets, market places and role-specific tool 
support. The core is to manage the interfaces between the different roles and the different assets such that all the 
participants can work independently of each other. The overall objective is to enable composability and compositionality 
of the assets in the ecosystem while adhering to the principle of {\em separation of roles}. Of course, this comes with 
structures like service-oriented software components with e.g. standardized state automatons and configuration interfaces 
to enable composition of control flows and data flows at design time (see fig. \ref{fig-components} and
\ref{fig-eclipse}) and even at runtime. The full details can be found in \cite{Schlegel2021}. 

Although these composition structures have been driven by the needs of service robot systems, they are in line 
with the needs and the concepts of many industry 4.0 scenarios. For example, the {\em digital data sheet} represents
all the information you need to know in order to predict how this asset behaves in your setting and to decide whether
it fits your needs. A digital data sheet is a model of the asset comprising all the information a user needs to know. It
is not suited for synthesis of an asset and does not disclose internals. In that sense, the various aspects of a digital 
data sheet of the service robotics ecosystem are in line with submodels of an AAS.

%
\section{AAS at Design Time}

\subsection{AAS for a Software Component at Design Time}

\begin{figure*}[htbp]
\centerline{\includegraphics[width=\textwidth]{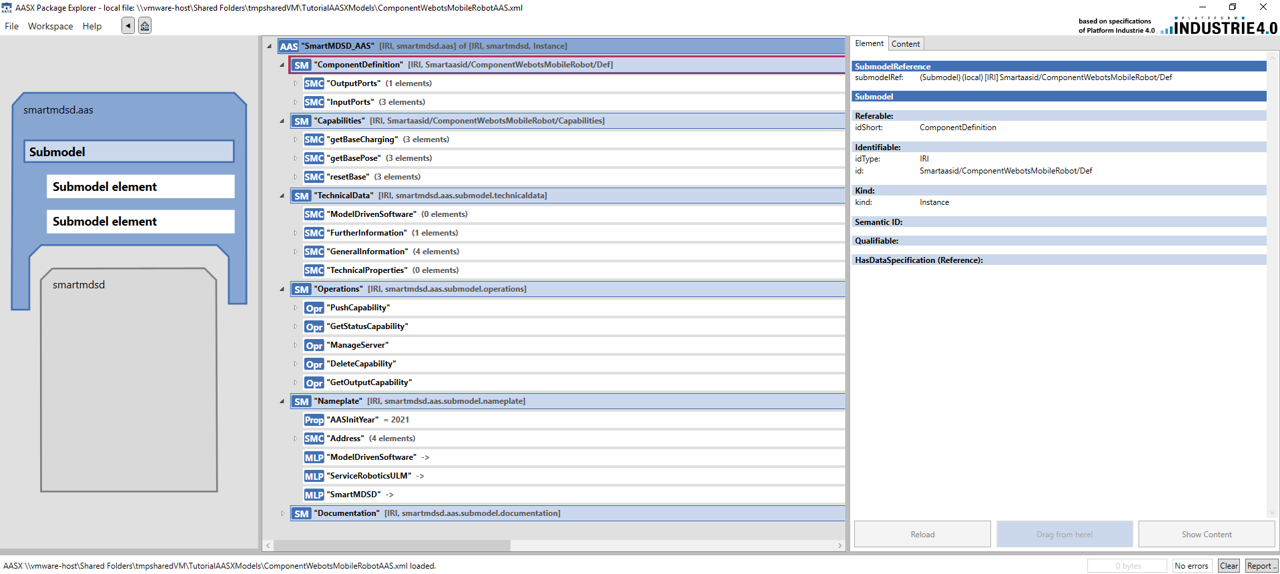}}
\caption{The XML file of the AAS of the software component for accessing the Webots robot simulator displayed via the AASX Package Explorer. 
The AAS of a software component comprises the submodels {\em ComponentDefinition}, {\em Capabilities}, {\em TechnicalData}, {\em Operations},
{\em Nameplate} and {\em Documentation}.}
\label{fig-aas-component}
\end{figure*}

Fig. \ref{fig-aas-component} shows an AAS of a software component. The semantics of the submodels and their elements is either defined by 
a pre-given standard for a submodel or it is given via the reference into the (domain-specific) models used by the model-driven toolchain {\sc SmartMDSD}.

The submodel {\em ComponentDefinition} gets directly filled from the model of the software component used in {\sc SmartMDSD} by the component 
developer to develop and build the software component. In the same way, the submodel {\em Capabilities} gets filled by the skills assigned by component
developers to the (sets of) software component(s) and available in the {\sc SmartMDSD} model(s) of the (sets of) software component(s).
The submodel {\em TechnicalData} comprises typical data sheet information in the form of name/value pairs like the license, the kind of environment 
in which the asset can be used etc. The submodel {\em Operations} comprises a standard set of operations to invoke the skills listed in 
the {\em Capabilities}. 

There exist different interaction patterns for AASs \cite{AAS-ReadingGuide}: file exchange is called type 1, access via an API is called type 2 and
peer-to-peer interaction is called type 3. We use an AAS for a (set of) software component(s) as type 1 AAS (download the AAS by the system builder
as AASX XML file to find, check, select software components) or as type 2 AAS (use an API to a local server of the downloaded AAS to view, extract and 
process relevant information for making your decisions).

\subsection{AAS for a System at Design Time}

\begin{figure*}[htbp]
\centerline{\includegraphics[width=\textwidth]{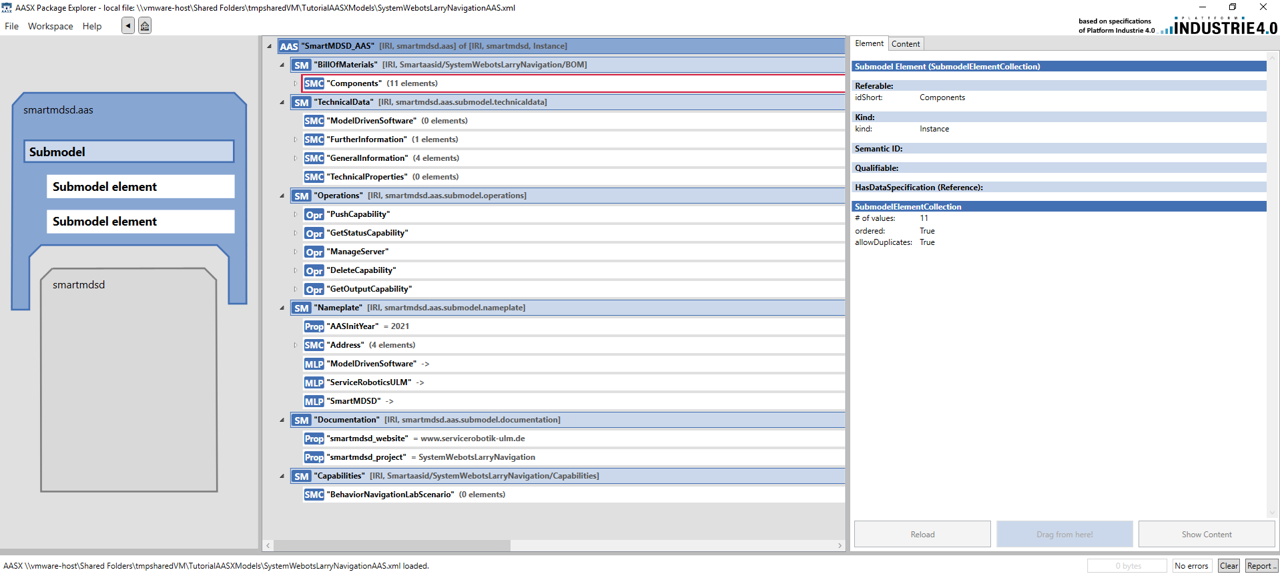}}
\caption{The XML file of the AAS of the robot system {\em Larry} deployed in the Webots robot simulator displayed via the AASX Package Explorer. 
The AAS of a service robotic system comprises the submodels {\em BillOfMaterials}, {\em TechnicalData}, {\em Operations}, {\em Nameplate}, 
{\em Documentation} and {\em Capabilities}.}
\label{fig-aas-system1}
\end{figure*}

Fig. \ref{fig-aas-system1} shows an AAS of a service robot system. Again, the submodels get filled from the models available via
{\sc SmartMDSD} when composing the system as system builder. The {\em BillOfMaterials} lists the components used in the system 
composition and the {\em Capabilities} is the complete list of all skills given by all the used software components and all the task plots 
added in the system composition step. However, there are good reasons to present only a subset of the skills and task plots to the 
outside and it might even be of interest to not give full insight into the bill of materials. Thus, in the future, we intend to exploit 
the access control mechanisms of AASs.

For a service robot system, we again use the AAS of the system as type 1 AAS (download of the AAS file) for finding, checking and 
selecting a service robot or as type 2 AAS (use an API to a local server of the downloaded AAS file to view, extract and process relevant
information for making your decision). The AAS of a service robot system can also hold accumulated operational data to support
decisions such as e.g. hours of operation, maintenance data and more.

%
\section{AAS at Runtime}

\subsection{AAS for Software Components at Runtime}

Right now, we do not use the AASs of the software components onboard of a service robot system at runtime to interact with
individual software components (parameterizing software components, invoking their skills, etc.). At runtime, the parameterizations
of software components, the selection of modes and the invocation of capabilities (in our terms {\em skills}) onboard the robot is
done without the indirection of the AAS as we have access to the first class citizen models of the software components which come
straight from the model-driven software development, composition and deployment processes.

\subsection{AAS for a System at Runtime}

\begin{figure*}[htbp]
\centerline{\includegraphics[width=\textwidth]{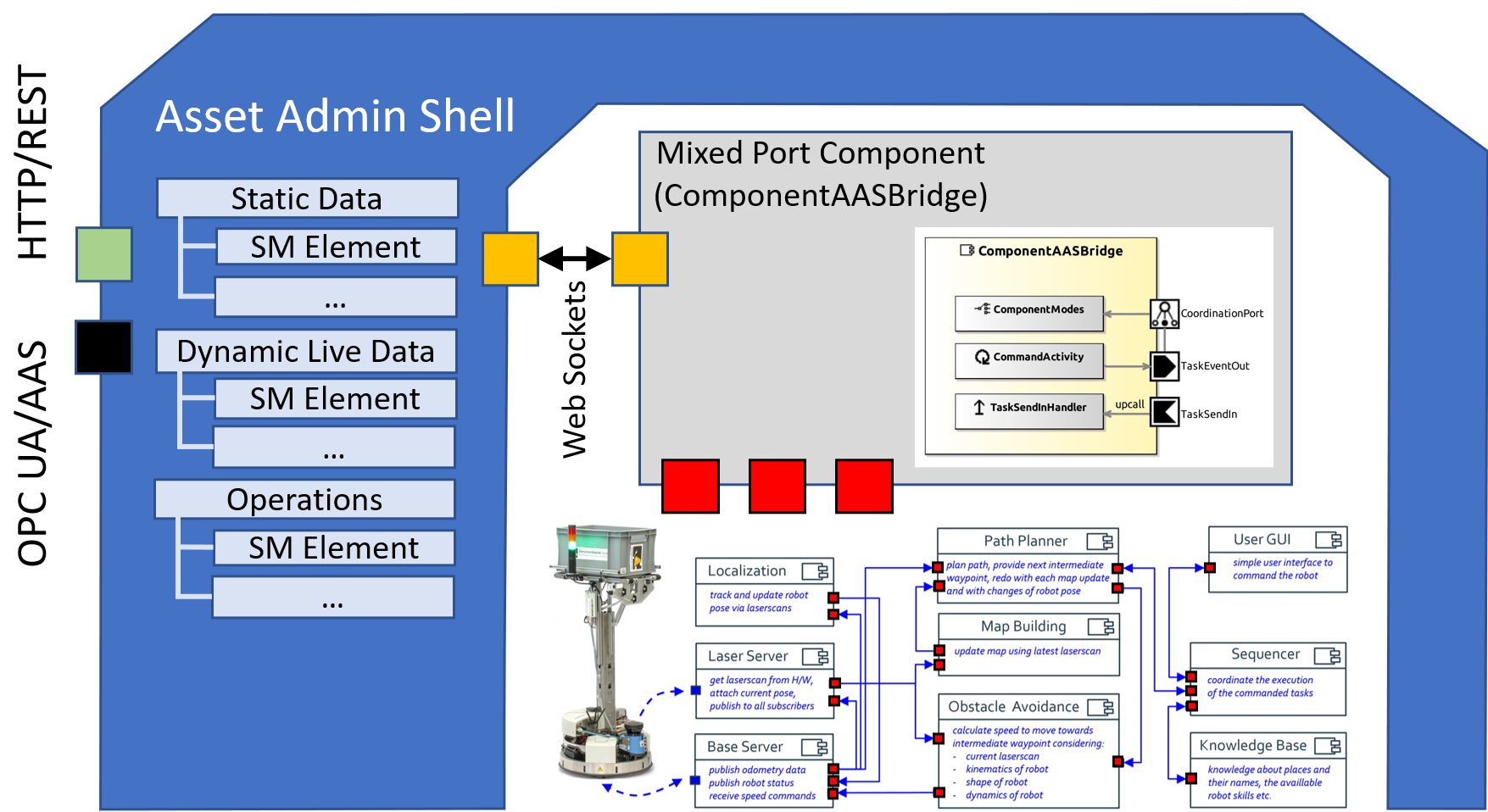}}
\caption{Using the AAS of a service robot system at runtime for commanding tasks and for reporting operational data. The outside 
access to the AAS conforms to the standardized structures of the AAS information models and be done via whatever technology is 
supported by the BaSyx SDK. Inside the AAS, we connect to a {\em Mixed Port Component} via Web Sockets (yellow ports). The AAS 
is implemented via the Java version of the BaSyx SDK while the software built with {\sc SmartMDSD} is C++. We could also use 
e.g. OPC UA for the yellow ports given that the BaSyx SDK fully covers the needed mechanisms. The yellow ports could be removed
in case the BaSyx C++ SDK is completed. The {\em Mixed Port Component} holds the business logic to interact with the composable 
software components of the robotic system (red ports). The advantage of the decoupling via the yellow ports is that we are free to
implement any business logic inside the {\em Mixed Port Component}.}
\label{fig-AAS}
\end{figure*}

Fig. \ref{fig-AAS} shows the type 2 use of an AAS of a service robotic system at runtime. The {\em Capabilities} can be used with 
{\em Operations}. Individual instances of a capability can be {\em pushed} (command it for execution) which returns an identifier 
that is then used for the {\em get status} operation (query the state of execution of the referenced command), the {\em get output} operation
(get its result) and the {\em delete} operation (remove the referenced command either before its execution or remove it after its completion and 
after you got its output).

A commanded capability can be in the states {\em pending} (execution not yet started), {\em executing} (in execution), 
{\em success} (execution successfully completed), {\em error} (completed but without success or with errors) or {\em deleted} (the identifier does 
not refer to a known commanded instance of a capability). 

We do not provide an individual operation per capability but a generic set of operations to be used with all the capabilities. You can also push several 
instances of the same capability as they all get unique identifiers. It is the robot which arranges the order of the execution by itself. A robot rejects a 
pushed command if e.g. the parameters or constraints do not fit or cannot be matched. It then runs immediately into an error for that command with 
further details in the result so that the reason for the reject can be checked.

This kind of command interface is quite typical for complex systems like service robots which have their own management of
resources and skills. We do not make an allocation of the service robot, select a particular mode, set the parameters for this very
mode and then invoke the related operation. We also do not let the service robot go into a bidding process for offered jobs (as would
be the case with a type 3 AAS). Instead, we enable every entitled entity to push a job to the robot for execution and have both, the 
immediate rejection by the robot without execution as well as the execution with either success or error and details of the outcome
reported via the results.

Another important part of the runtime use of the AAS is to collect all the operational data such as kilometers travelled, success
rates of task completion, time needed for tasks etc. Basically, the AAS allows for a standardized interface to access this information,
to monitor a service robot, to assess its performance and to command it.

%
\section{Conclusion and Future Work}

An AAS for a service robot has valuable use-cases at design time and at runtime. It is not meant to have all the communication 
inside a service robot organized via an AAS or by only using OPC UA. It is about the right granularity of providing a standardized access 
to complex machines and let them interact via standardized means and protocols. It got obvious that a type 1 AAS can serve as a 
digital data sheet whereas a type 2 AAS with a skill-based approach fits best for embedding service robots into Industry 4.0 settings
\cite{SIDORENKO2021191}.

Already now, one can use the virtual machine and the related tutorials to get an impression and to play around with our AASs for
service robotic systems.

The next steps are extensions of the skill-based interface to better include a kind of bidding in the commanding so that we do not
just end up in an error state with {\em rejected} as explanation. Thereto, we will extend both, the set of constraints and parameters 
that can be given with a {\em push} and enrich the information that is returned in acknowledging respectively rejecting a given
command.

It is important to note that we can easily adjust our submodel templates to upcoming standardizations as we use a model-driven
approach with DSLs (domain specific languages) to map from our {\sc SmartMDSD} models to AAS submodel templates.

%
\section*{Resources}

\noindent Ready-to-run downloads and tutorials:
\begin{itemize}
\item \url{https://wiki.servicerobotik-ulm.de/virtual-machine}
\item \url{https://wiki.servicerobotik-ulm.de/tutorials:start}
\end{itemize}

\noindent AAS for components and systems:
\begin{itemize}
\item \url{https://wiki.servicerobotik-ulm.de/tutorials:start#adding_aas_to_components_and_systems}
\item \url{https://wiki.servicerobotik-ulm.de/tutorials:start#interacting_with_the_aas}
\end{itemize}

\noindent A bigger picture of the kind of service robotic systems and applications we are addressing:
\begin{itemize}
\item \url{https://www.youtube.com/user/RoboticsAtHsUlm/videos}
\end{itemize}

%
\section*{Open Access}

This paper is licensed under the terms of the Creative Commons Attribution 4.0 International License (\url{http://creativecommons.org/licenses/by/4.0/}), which permits use, sharing, adaptation, distribution and reproduction in any medium or format, as long as you give appropriate credit to the original author(s) and the source, provide a link to the Creative Commons license and indicate if changes were made.
\begin{figure}[htb]
\centerline{\includegraphics[height=2cm]{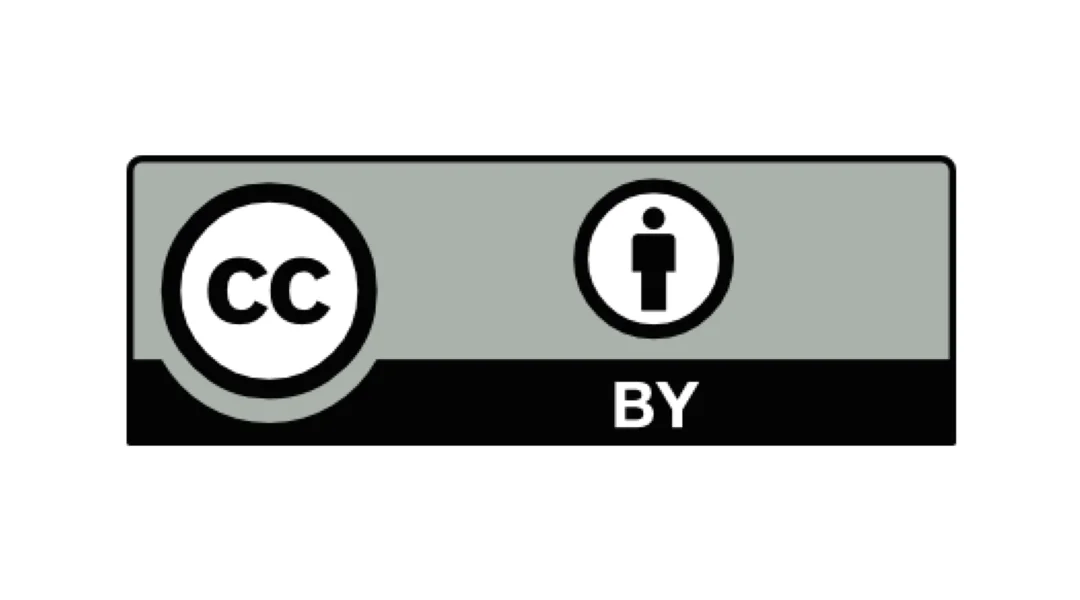}}
\end{figure}

%
\bibliographystyle{IEEEtran}
\bibliography{AAS-SmartMDSD-2022-WorkInProgress}

\end{document}